\documentclass[letterpaper]{article}
\usepackage{aaai}
\usepackage{times}
\usepackage{helvet}
\usepackage{courier}
\usepackage{graphicx}
\usepackage{subfigure}
\usepackage{bbm}
\usepackage{todonotes}
\usepackage{amsfonts} 
\usepackage{amsmath}
\usepackage{booktabs}
\usepackage{array}
\usepackage{multirow}
\usepackage{arydshln}
\newcolumntype{L}[1]{>{\raggedright\let\newline\\\arraybackslash\hspace{0pt}}m{#1}}
\newcolumntype{C}[1]{>{\centering\let\newline\\\arraybackslash\hspace{0pt}}m{#1}}
\newcolumntype{R}[1]{>{\raggedleft\let\newline\\\arraybackslash\hspace{0pt}}m{#1}}

\DeclareMathOperator*{\argmin}{arg\,min}
\newcommand{\V}[1]{\mathbf{#1}}

\begin{document}
%
\nocopyright  

\title{Outcome-Guided Counterfactuals for Reinforcement Learning Agents from a Jointly Trained Generative Latent Space}
\author{Eric Yeh, Pedro Sequeira, Jesse Hostetler, Melinda Gervasio\\
SRI International\\
333 Ravenswood Avenue\\
Menlo Park, California 94502\\
}
\maketitle
\begin{abstract}
\begin{quote}
We present a novel generative method for producing unseen and plausible counterfactual examples for reinforcement learning (RL) agents based upon outcome variables that characterize agent behavior.  Our approach uses a variational autoencoder to train a latent space that jointly encodes information about the observations and outcome variables pertaining to an agent's behavior.  Counterfactuals are generated using traversals in this latent space, via gradient-driven updates as well as latent interpolations against cases drawn from a pool of examples.  These include updates to raise the likelihood of generated examples, which improves the plausibility of generated counterfactuals.  From experiments in three RL environments, we show that these methods produce counterfactuals that are more plausible and proximal to their queries compared to purely outcome-driven or case-based baselines.  Finally, we show that a latent jointly trained to reconstruct both the input observations and behavioral outcome variables produces higher-quality counterfactuals over latents trained solely to reconstruct the observation inputs.  
\end{quote}
\end{abstract}




\section{Introduction}
\noindent Consider a scenario where a human user needs to decide whether a self-driving vehicle will be behave as expected when driving down a busy street.  Policy characterization techniques, in particular, interestingness variables \cite{sequeirainterestingness2019}, capture key meaningful situations for an agent's policy and identify when an agent's behavior may change. However, understanding what factors affect these situations and how they affect them is also important for human decision-makers. Per our use case, the user wishes to determine what could impact \emph{riskiness}, an interestingness variable that measures the agent's perception of how consequential its actions can be.  Identifying what can increase or decrease this value can help the user determine when the agent is entering a regime where it may be less tolerant of small deviations and mistakes when executing its policy, or if the agent indeed is performing a correct assessment of a scene.

A common approach is to explain the agent's behavior in terms of the features it deems most important to its riskiness assessment---e.g., through attention maps that highlight the portions of the street immediately in front of the vehicle.  While these can be effective for debugging machine learning systems, they can be difficult for non-experts to understand \cite{wachtergdpr2017}.  On the other hand, \emph{counterfactuals}---i.e., contrastive, example-based explanations that \emph{change} specific aspects of the environment to arrive at a different outcome---are more aligned with how humans develop an actionable understanding of autonomous systems \cite{millerxaisocial2017}.  Several explainable AI studies have also shown them to be an effective tool for conveying explanations about AI systems \cite{byrnehuman2019}.  Following our example, adding jaywalkers to the scene increases riskiness, assuring the user that the agent recognizes the danger to the pedestrians and thus is aligned with human expectation.   

There are two main challenges to generating counterfactuals for these systems. The first is that not all of the possible inputs or reasoning components of the agent may be available to the inspector---training data and environments may not be available, or the platform itself may not be amenable to such experimentation.   The second is that counterfactuals need to be plausible and meaningful to the human.  While creating realistic and relevant examples by perturbing features is feasible for discrete, low-dimensional data, this becomes more difficult for complex, high-dimensional inputs, such as visual and spatial data \cite{wachtergdpr2017}.  

To address the first challenge, we present a novel generative model that focuses on ``black-box'' analyses of reinforcement learning (RL) agents that trains on observational data.  Using a corpus of performed trajectories and corresponding outcome variables, we train a  variational autoencoder \cite{kingmavae2013} modified to jointly reconstruct the agent's observations and predict outcome variables that characterize its behavior.  By sampling from this latent space, we can identify when a selected variable has changed significantly and generate the observation corresponding to the modified latent vector---i.e., its \emph{counterfactual}.  We demonstrate two methods for doing this. The first uses interpolations in this latent space towards case-based examples.  The second generates counterfactuals using iterative gradient-guided traversals in this latent space, following work in Plug-and-Play models to modify generative models \cite{DBLP:journals/corr/NguyenYBDC16}.   

For the second challenge, we demonstrate how a gradient-based \emph{plausibility adjustment} to raise the likelihood of generated examples improves the quality of produced counterfactuals.  As numerous studies using generative models have shown, samples drawn even from generatively trained latent spaces are not guaranteed to produce sufficiently plausible examples, i.e., examples that are within the expected distribution of the input data \cite{malisnick_dgn_know_2018}. 
We show that performing this adjustment can reduce the number of anomalous counterfactuals and their degree of anomaly.

Finally, we present empirical evidence that training a model that jointly reconstructs both the input and the outcome variables of interest both improves the number of valid counterfactuals and their quality, in contrast to similar non-jointly-trained generative approaches.

We summarize our contributions as follows,
\begin{itemize}
    \item A flexible Plug-and-Play framework for gradient-driven traversal over a latent space that can incorporate multiple types of constraints to produce unseen and plausible counterfactuals.  We show that this can be used to improve case-based and gradient-derived counterfactuals.  This can also target multiple outcome variables at once, as well as numeric and categorical outcomes.
    \item Demonstrate the need for ensuring the plausibility of generated counterfactuals, and present a gradient-driven approach to improve plausibility.  
    \item A novel approach leveraging a jointly trained latent space and a demonstration of its importance for improving the quality of generated counterfactuals.
\end{itemize}

These approaches are detailed in their own sections, with accompanying experimental results over three reinforcement learning environments.  We note that while our approach improves counterfactual quality along several measures, an extrinsic evaluation of counterfactual usefulness, in particular for interestingness analysis, is still in development.

\section{Related Work}
Automated counterfactual generation has been explored by numerous communities.  One line of work has focused on case-based approaches, which produces counterfactuals by drawing from a library of observed instances.  Suitable counterfactuals are either selected from this library or used as a basis for synthesizing counterfactuals.  Identifying the Nearest Unlike Neighbor (NUN) is perhaps the simplest approach, and also ensures plausibility \cite{keanegoodcounterfactuals2020}.  The NUN is the instance that is closest to the query, based on a proximity measure such as edit distance, whose outcome variable of interest has changed, e.g., has a different label associated with it.  However, in many cases a NUN cannot be identified, as no example in the library is close enough to the query such that the value of the corresponding outcome variable has changed sufficiently.  

The approach given in \cite{keanegoodcounterfactuals2020} identifies good counterfactual pairs in the data and uses them as templates for feature transformation. Given a query, if a NUN cannot be found, the closest pair is selected and the feature-level differences between the examples in that pair are used to transform the query into the target.  Another approach uses feature-relevance methods like SHAP \cite{lundbergshap2017} to tailor a feature edit schedule for converting the query into a counterfactual \cite{DBLP:journals/corr/abs-2109-05800}.  In this body of work, an imitating function is trained to predict the outcome variable given the instance features.  This is then used to ensure that the synthesized counterfactual does indeed change the outcome; we use this ``black-box'' approach to validate our counterfactuals as well.
We note that most of this work focuses on tabular data, where data is organized into discrete features (e.g., columns) and copying values has less risk of creating an implausible instance than with less structured data, such as images.  

Another body of work stems from the adversarial AI community, where input perturbation and sampling-based methods are used to generate counterfactuals \cite{Szegedy14intriguingproperties}. In particular, a Generative Adversarial Network (GAN) was used to generate counterfactual scenes for non-tabular data \cite{DBLP:journals/corr/abs-2101-12446}.  However, sampling for suitable solutions can be time-consuming, whereas gradient- or example-directed traversals can be more efficient.
\begin{figure*}
    \includegraphics{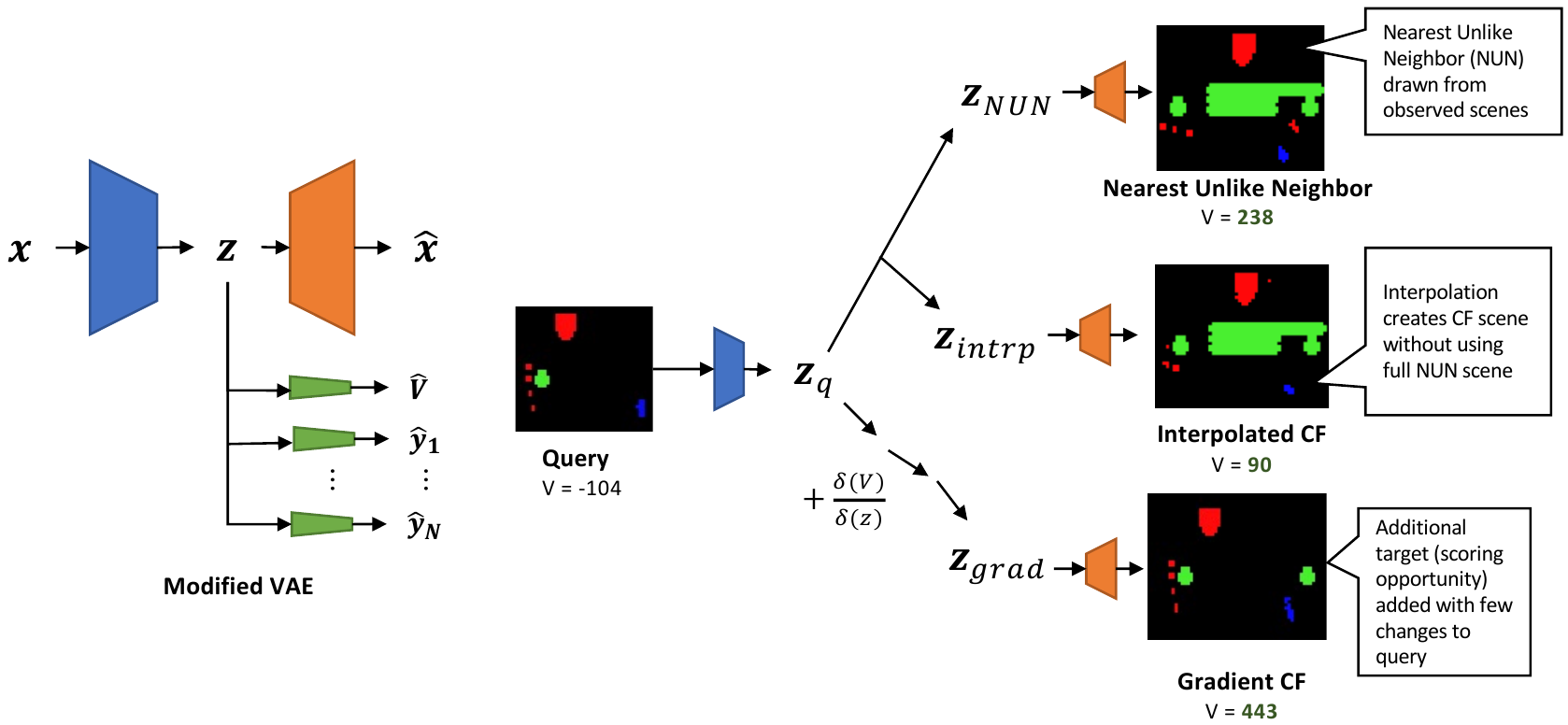}
    \caption{A variational autoencoder is trained to both reconstruct the input and predict several outcome variables such as the agent's value function and other behavior-related variables of interest (left).  As an example, given a \emph{low-value} query instance in the StarCraft II domain encoded as spatial feature layers using PySC2\footnote{https://github.com/deepmind/pysc2} (agent units are shown in are blue; enemies, red, and assets that can be captured, green), the goal is to obtain a \emph{higher-valued} counterfactual.  The query instance is encoded to its latent ($z_q$) and counterfactuals (CF) are obtained by three methods (right): identifying the Nearest Unlike Neighbor from data (NUN, top) and then doing a partial latent interpolation to the NUN (middle), or using the gradient information to generate the example (bottom).  In this example, adding an additional target (green circle) for the agent raises the value estimate.  The interpolation and gradient methods produce alternatives with fewer defending forces and placement of obstacles than the NUN.}
    \label{fig:overview}
\end{figure*}

Gradient descent in the feature space towards the desired outcome has been used for tabular data \cite{DBLP:journals/corr/abs-1905-07697} and, in particular, in the adversarial machine learning community for generating counterfactuals over image data \cite{DBLP:journals/corr/abs-1906-10671,DBLP:journals/corr/abs-2101-06930}.  However, these approaches run the risk of generating adversarial counterfactuals, such as shifting a minimal set of pixels, that may be imperceptible to human users and have low utility for understanding the model.

Our approach is motivated by the Plug-and-Play approach, which uses gradient-derived signals from discriminators to iteratively shift the latent of generative models to produce outputs with the desired characteristics, and Plug-and-Play language models (PPLM), which adds gradient adjustments to increase the probability of the generated instances \cite{DBLP:journals/corr/NguyenYBDC16,DBLP:journals/corr/abs-1912-02164}.  Iterative gradient-based adjustment of a latent space was also explored in the xGEMs approach \cite{joshixgems2018}, albeit in that case, the latent was not trained to include outcome information and reconstructions were not adjusted for plausibility.  A similar approach was taken for attribute-based perturbation in the latent space of a conditional variational autoencoder (C-VAE) for counterfactual generation \cite{DBLP:journals/corr/abs-2101-06930}.  This approach, closest to ours, used an outcome embedding trained separately from the latent for the features, and both are concatenated and used as inputs to the decoder.  In contrast, however, our approach focuses on a latent that jointly encodes the data and outcome variables, with predictions made from the latent itself.  In addition, these approaches have neither explicitly measured nor attempted to ensure plausibility.  Works that do address this have focused on computer-vision measures of image quality, such as Fr\'{e}chet Inception Divergence \cite{FID} which uses statistics over feature-activations in a network to act as a form of perceptual distance, and is used to identify unrealistic artifacts such as blurry images.  As multiple studies have shown, generative models are not necessarily guaranteed to produce samples that are plausible from the in-distribution set, and taking measures to avoid anomalous examples is required \cite{malisnick_dgn_know_2018}.  In addition, these approaches appear to mostly avoid the issue of proximity as well, and focus only on whether the generated counterfactual actually changes the desired outcome.


\section{Preliminaries}
We now describe preliminaries for this work, starting with background on the generative model followed by how counterfactual queries are formulated.  We then describe our metrics for counterfactual effectiveness and how they are implemented in our experiments.   

\subsection{Variational Autoencoders}
Variational autoencoders (VAEs) are a type of probabilistic generative model that encodes a high-dimensional input $\V{x}$ into a lower-dimensional latent representation $\V{z}$ from which the original input can be approximately reconstructed \cite{kingmavae2013}. The encoder module of the VAE maps the input to its latent representation, $\V{z} = enc(\V{x})$, and the decoder reconstructs the input, $\V{x} \approx dec(\V{z})$.
The latent encoding is regularized by penalizing the KL divergence from a prior distribution $q(\V{z})$ (typically a standard Gaussian) to the conditional distribution $q(\V{z}|\V{x})$ induced by the encoder. 
The VAE loss is given by
\[ \mathcal{L}(\V{x}) = \mathbb{E}_{q(\V{z}|\V{x})} \log p(\V{x}|\V{z}) - D_{KL}(q(\V{z}) || p(\V{z}|\V{x})), \]
where the decoder likelihood $p(\V{x}|\V{z})$ is typically implicit from a reconstruction loss, such as the MSE $||\V{x} - dec(\V{z})||^2$. It can be shown \cite{kingmavae2013} that the VAE loss is a lower bound on the data likelihood, $\mathcal{L}(x) \leq \log p(\V{x})$. Because the encoder distribution $q(\V{z}|\V{x})$ approximates the known prior distribution $q(\V{x})$, samples from the input space can be generated by drawing $\V{z} \sim q(\V{z})$ and passing the result through the decoder.


\subsection{Counterfactual Generation with VAEs}
Let $M$ denote the model whose behavior we wish to explore using counterfactual analysis. Given a query input $\V{x}_q$, we want to generate a counterfactual input $\V{x}_c$ that is ``related'' to $\V{x}$ but for which $M$ would behave differently. We quantify the behavior of $M$ with a vector of outcome variables $\V{y} = (y^{(i)})$, $i \in \{1, \ldots, N\}$. When $M$ is a reinforcement learning agent, for example, $\V{y}$ might include the value achieved by the agent, secondary performance measures such as the time needed to reach a goal, and/or categorical measures like whether the agent violated certain constraints.

Our approach to counterfactual generation is based on perturbing the latent representation $\V{z}_q$ of the query input to create a counterfactual latent representation $\V{z}_c$, then decoding $\V{z}_c$ to obtain a counterfactual input $\V{x}_c$, exploiting the ability of VAEs to learn a latent representation space with meaningful axes of variation \cite{higgins2016beta,klys2018learning}.
We extend the basic VAE model to reconstruct both the input $\V{x}$ and the outcome variables $\V{y}$ from the latent representation $\V{z}$, using a separate predictor for each outcome variable, $y^{(i)} = \sigma_i(\V{z})$ (Figure~\ref{fig:overview}, Left). Our intent is to cause the latent representation to encode information about the relationship between the input and the outcome variables, so that traversing the latent space will produce inputs that result in different outcomes.
This also provides a trained predictor that can indicate when an example meets the counterfactual outcome criteria.  This use of a trained proxy to determine if the outcome is met is a common practice in the counterfactual generation literature from observational data \cite{keanegoodcounterfactuals2020}.



We say that a counterfactual is \emph{valid} if it achieves a desired change in the outcome variables. We define a validity predicate $\kappa_{i,s,\epsilon}$ that indicates whether the $i$th outcome variable was changed appropriately, given by
\begin{equation*}
\kappa_{i,s,\epsilon} = \left\{
    \begin{array}{ll}
        \mathbb{I}(y_c^{(i)} \neq y_q^{(i)}) & \text{if } y^{(i)} \text{ is categorical} \\
        \mathbb{I}(y_c^{(i)} - y_q^{(i)} \geq s\epsilon) & \text{if } y^{(i)} \text{ is numeric}
    \end{array} \right.,
\end{equation*}
where $s \in \{-1,1\}$ is the desired sign of the difference between numeric variables, and $\epsilon$ is the desired size of the difference. For brevity, we shorten the validity predicate to $\kappa_i$ for the rest of this paper. While our experiments consider only a single criterion (a single $i$), our approach can be extended easily to multiple criteria.


\subsection{Counterfactual Quality Measures}
For this work, we evaluate counterfactual generation methods along the following measures:  
\begin{itemize}
    \item \textbf{Proximity}: How different a generated counterfactual is from its query.
    \item \textbf{Plausibility}: Whether the counterfactual is something one would expect to observe in the domain of interest.
    \item \textbf{Validity}: Whether the counterfactual satisfies the counterfactual criterion  $\kappa_i$.
\end{itemize}

Keeping the counterfactual similar to the original instance is important for understanding the relation between the features and the outcome variables and \emph{proximity} is commonly measured through a variety of feature-level edit distance metrics, with counterfactuals having a \emph{sparser} set of differences from the query being better. \emph{Plausibility} is particularly important for systems that synthesize counterfactuals, as anomalous or implausible examples may be discounted by users \cite{keanegoodcounterfactuals2020,millerxaisocial2017}. Is is often measured by how likely the counterfactual is to be drawn from the actual data,
Finally, \emph{validity} is necessary since automatically generated counterfactuals may not actually meet the counterfactual criterion $\kappa_i$.  

\begin{figure*}
    \centering
        \includegraphics[scale=0.7]{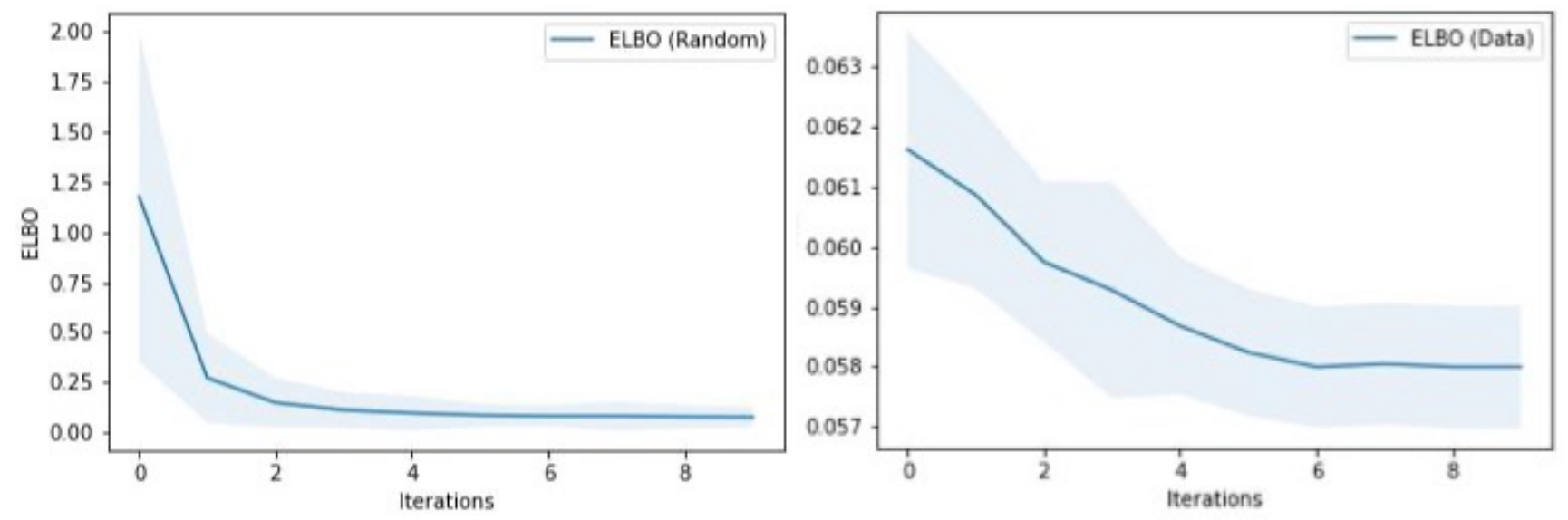}
    \caption{Evidence lower bound (ELBO) loss against number of round trips for the input, starting from data drawn instances (top) or randomly sampled (bottom).  The ELBO lower bounds the log-likelihood of the input, thus as the loss decreases the likelihood of the input increases.}
    \label{fig:roundtrip_elbo}
\end{figure*}
We measure the inverse of proximity of a counterfactual to its query via an \emph{observational difference} score.  For categorical features, this is computed as a feature edit distance, summed over the number of label changes to convert between observations.  For numeric features, we use the absolute score difference, normalized to 0-1.  Formally, for an index of all features, the score $i \in I$, $\mathrm{odiff}(x^1, x^2)$ is computed as follows:
\begin{equation*}
\mathrm{odiff}(\V{y}_1, \V{y}_2) = \sum_i \left\{
    \begin{array}{@{}ll@{}}
        \mathbb{I}(y_1^{(i)} \neq y_2^{(i)}) &  \text{if } i \text{ is categorical} \\
        \\
        \frac{|y_1^{(i)} - y_2^{(i)}|}{W^{(i)}} & \text{if } i \text{ is numeric}
    \end{array} \right.,
\end{equation*}
where $W^{(i)}$ is the interval width that normalizes the value.

Finally, there is no guarantee that a given method can produce a \emph{valid} counterfactual, one that satisfies the criterion $\kappa_i$.  We grade each counterfactual generation method by the fraction of queries for which it was able to produce a valid counterfactual, that is $\frac{1}{N} \sum^N_{i=1} \kappa_i$ for $N$ queries.  

\subsection{Plausibility}
For a given latent $\V{z}$, we assess its plausibility by measuring how anomalous its decoding is.  Following the observation that autoencoders act to denoise anomalous inputs \cite{NIPS1992_b7bb35b9}, we construct an anomaly score using the VAE by measuring the observational difference between the decoded instance and its reconstruction,

\begin{equation*}
    \text{anom}(\V{z}) = \text{odiff}(dec(\V{z}), dec(enc(dec(\V{z}))))
\end{equation*}  

We use this anomaly score to measure (the inverse of) plausibility in our experiments.\footnote{We did experiment with One-Class SVMs, but performance on a SC2 Assault scene labeled for anomalous scenes was poor in comparison with the autoencoder approach. }

As stated earlier, ensuring generated counterfactuals are plausible to the end user is important, as implausible or unrealistic elements may reduce their effectiveness. If we knew the likelihood function $p(\V{x})$, we could improve plausibility by performing a \emph{plausibility adjustment} to the latent representation to increase the likelihood of its decoding, i.e., $\V{z} \leftarrow \V{z} + \nabla_{\V{z}} p(dec(\V{z}))$. Inspired by the hypothesis that autoencoders denoise their inputs, we approximate this gradient with the reconstruction loss between the current latent's reconstruction and that scene's reconstruction using the same model,  
\begin{equation*}
    \nabla_{\V{z}} p(dec(\V{z})) \approx -\nabla_{\V{z}} ||dec(\V{z}) - dec(enc(dec(\V{z})))||
\end{equation*}

We verified the appropriateness of using this approach to increase the plausibility of generated counterfactuals by having our VAE repeatedly encode and decode its own reconstructions.  Figure~\ref{fig:roundtrip_elbo} shows the mean and standard deviation of the ELBO loss\footnote{ELBO loss drawn from our training setup, with a KL scaling term of $\beta=10^{-5}$.} of the input at each step of the recurrence.  The top figure shows the curve for 1000 scenes sampled from our Starcraft II minigame data (described in the following sections).  The plot on the bottom repeats this procedure but with 1000 randomly sampled latents, drawn from the unit normal with random scale from 0-10.  The ELBO (in non-loss form) lower bounds the model's log-likelihood of the data.  Indeed, in both cases we see a decrease in ELBO loss, or increase in instance likelihood, with the greatest impact at the first step.  

We note that while likelihoods from deep generative models may not be sufficiently calibrated for outlier detection \cite{wang_outlier_dgm_2020}, we are not attempting to estimate a distribution, but instead are looking to increase the likelihood---hence the plausibility---of the reconstructions during counterfactual generation.  Indeed, recent work has found that measures of likelihood from deep networks are more congruent with how familiar a model is with the features of a scene \cite{dietterich_familiarity_2022}.  This matches our use case, as we are relying on the model's tendency to reconstruct using elements it is more familiar with.

\section{Counterfactual Methods}

Figure \ref{fig:overview} (right) illustrates our counterfactual generation architecture featuring the three methods in this work.  The first draws a suitable example, the Nearest Unlike Neighbor (NUN), from previously observed examples.  The second uses a traversal in the jointly trained latent space between the query and the NUN example.  By stopping the traversal when the counterfactual criterion is met, this interpolated counterfactual is more proximal to the query (requires fewer feature edits).  The third approach uses gradient information provided by the outcome predictors to perform a directed search in the latent space.  

\subsection{Nearest Unlike Neighbors}
The most reliable way to obtain a valid and plausible counterfactual is to draw it from a library of observed examples. The Nearest Unlike Neighbor (NUN) is an instance that is similar to the query, but has an outcome that meets the counterfactual criterion \cite{keanegoodcounterfactuals2020}.  In our experiments, we draw the NUN $\langle \mathbf{x}_q, \mathbf{y}_{NUN} \rangle$ from the VAE's training instances, minimizing the observational distance while meeting the counterfactual criterion,
\begin{equation*}
    \V{x}_{NUN} = \argmin_{\V{x}_c} \text{odiff}(x_q, x_c) \quad \text{s.t. } \kappa_{i} = 1.
\end{equation*}

\subsection{Latent Interpolation}
While drawing NUNs can ensure plausibility, there is no guarantee of proximity: There may be no instances that are sufficiently similar to the query. Several studies have generated counterfactuals for tabular data by interpolating between the query and the NUN \cite{keanegoodcounterfactuals2020,DBLP:journals/corr/abs-2109-05800}.  As we are using a generative model, we can perform a similar interpolation in the latent space by interpolating linearly between the latent encodings of the query $\V{z}_q$ and 
the NUN $\V{z}_{NUN}$ to obtain the interpolated latent representation $\V{z}_{\iota}$.  The scaling factor $\alpha$ is sampled from 0 to 1, set to the first point where the counterfactual criterion is first satisfied, i.e., $\kappa_i(\sigma_i(\V{z}_{\iota})) = 1$.  
If $\alpha=1$, we consider the interpolation to have failed to produce a valid counterfactual, as the result is the NUN.  
If a point was found, we then update $\V{z}_{\iota}$ with a plausibility adjustment, with the magnitude $\lambda$ selected by a grid search along the unit direction of the gradient for the point with the lowest anomaly score.

\begin{equation*}
    \begin{split}
    \V{z}_{\iota} & = \alpha (\V{z}_{NUN} - \V{z}_q) + (1-\alpha)\V{z}_q,\quad \alpha \in [0, 1] \\
    \V{z_{\iota}} & = \V{z}_{\iota} + \lambda \nabla_{\V{z}} p(dec(\V{z_{\iota}}))
    \end{split}
\end{equation*}


\subsection{Iterative Gradient Updates}
Instead of relying on interpolating toward a concrete example, we can simply follow the gradient signal from the desired outcome predictor to shift it in the desired direction.  We then apply a plausibility adjustment to shift the latent to a higher likelihood state.
\begin{equation*}
 \begin{split}
  \V{z} & = \V{z} + s \lambda_1 \nabla_{\V{z}} y^{(i)} \\
  \V{z} & = \V{z} + \lambda_2 \nabla_{\V{z}} p(dec(\V{z})).
 \end{split}
\end{equation*}

Where $s \in \{-1,1\}$ is the desired sign of the change, and scaling terms $\lambda_1, \lambda_2$.\footnote{Set to $\lambda_1=5, \lambda_2=1$, tuned over the training set.}  We iterate this update until the counterfactual criterion $\kappa_i(\sigma_i(z))$ is satisfied or a maximum number of steps is reached \footnote{This was arbitrarily set to 1000 in our experiments.}.  We note that the gradient update over a latent space trained only for reconstruction, without the plausibility adjustment, is equivalent xGEMs \cite{joshixgems2018} and similar methods in the literature.  
 \begin{figure*}
    \centering
    \includegraphics[scale=0.68]{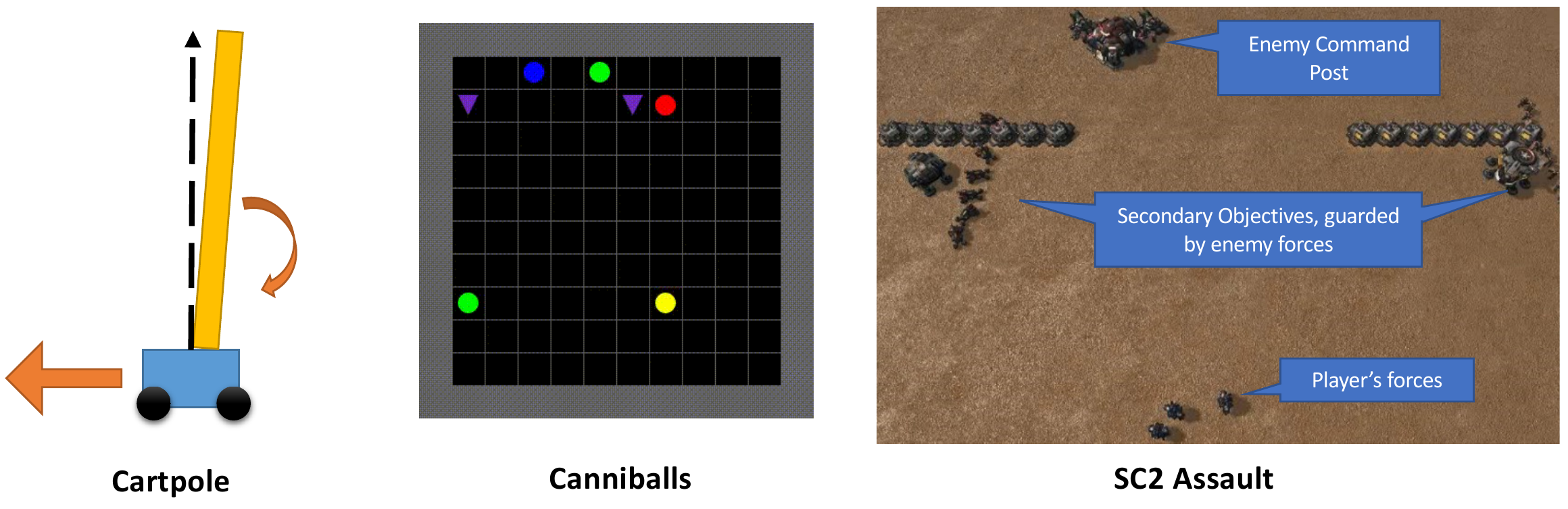}
    \caption{Environments used in counterfactual generation experiments.  }
    \label{fig:environments}
\end{figure*}
\section{Experiment}
We now describe our experimental setup.  We describe the three environments used, along with details on the model, RL training, and counterfactual query setup.  We follow with results detailing counterfactual quality across three environments, and show the effect of the plausibility adjustment and its impact against a concrete number definition of anomalous counterfactuals.  When then show how the joint training helps with improving counterfactual quality in comparison with a reconstruction-only latent.


\subsection{Environments}
In order to assess generalizability of our counterfactual methods, we conducted experiments in three different reinforcement learning environments: Cartpole \cite{brockman2016openai}, Canniballs \cite{cameleon}, and a custom minigame in the StarCraft II Learning Environment 
\cite{DBLP:journals/corr/abs-1708-04782}.  Figure \ref{fig:environments} illustrates the environments.

Cartpole (left) is a two-dimensional physics simulation, where the agent has to balance a pole on a cart by moving left and right.  Reward is given for each timestep the pole remains upright and balanced, with episodes ending when the pole falls over or the cart veers too far from its origin.  Observed state consist of four continuous parameters: cart velocity and position, pole angle and angular velocity.

Canniballs (center) \cite{cameleon} is a gridworld game designed to exercise multiple subgoals in a highly stochastic environment.  The player controls the red ball, and reward is earned for consuming weaker entities in the game, with a penalty applied for stalling or being consumed.  Episodes end when the player is consumed or after a fixed number of steps.  All game entities have a strength level, including the player, who can only consume entities weaker than itself. Strength is built up by consuming different entity types (colored balls and triangles), where balls have their own behavior, such as random movement, bouncing across the field, or chasing the player.  Observations are in the form of a set of categorical spatial feature layers.

StarCraft II\footnote{https://StarCraft2.com} (right) is a multiplayer real-time strategy game that features a variety of unit and building types.  Each unit type has strengths and weaknesses, and part of the strategy is to employ the best unit to win the game.  Buildings provide unique capabilities and can be destroyed or seized.  For our experiments, we developed a custom scenario designed to exercise complex decision-making.  The agent takes the part of one of the players, and is rewarded for destroying enemy forces, seizing secondary objectives, and destroying the enemy's command post.  Capturing a secondary objective provides the player with reinforcements, which can be used to avoid obstacles.  The observation space is spatial, but contains multiple layers containing both numerical and categorical data, and is significantly more complex than the two other environments.  For the remainder of this paper, we will refer to this scenario as \emph{SC2 Assault}.

\subsection{Reinforcement Learning}
Both Cartpole and Canniballs were trained using the RLLib framework \cite{DBLP:journals/corr/abs-1712-09381}.  For the SC2 Assault agent, we used a V-trace \cite{DBLP:journals/corr/abs-1802-01561} agent trained using the Reaver toolkit \cite{reaver}.  This was based in the StarCraft II Learning Environment via the PySC2 interface \cite{DBLP:journals/corr/abs-1708-04782}, using a subset of the full action set that is focused on movement and attacks for each type of unit.  Having trained the RL agents, we produced $1000$ episodes for each environment using the trained policy. This resulted in $189674$ frames for Cartpole, $136671$ for Canniballs, and $213407$ for SC2 Assault.

For the outcome variables used to form counterfactual queries, we based our approach on the concept of interestingness elements \cite{sequeirainterestingness2019,sequeirainterestingness2020}, corresponding to numeric measures that allow highlighting meaningful and potentially explanatory situations as an RL agent interacts with its environment.  Each measure is derived from data representing the agent's internal state, such as the value function estimate, $V$, the action value function $Q$ (depending on the architecture), the action distribution, and others.  For these experiments, we used used the following interestingness variables as outcomes for exploring counterfactuals:

\begin{itemize}
    \item \textbf{Value}: The value function estimate, measuring the expected discounted cumulative reward at any given state.
    \item \textbf{Confidence}: The action execution certainty of the agent, where we use a measure of statistical dispersion that relies on the entropy of the policy's action distribution.
    \item \textbf{Riskiness}: The margin between highest- and lowest-valued outcomes from taking an action, representing the perceived tolerance for mistakes in the environment.
\end{itemize}

\subsection{Model}
We now describe the VAE used to construct the surrogate model from agent trajectories.  For the Cartpole agent, we used MLP encoders and decoders over the vector.  Canniballs and SC2 Assault use spatial features, for which we used a convolutional architecture encoder and decoder.  
The variational autoencoder was also tuned with a fixed beta schedule going from $\beta=0$ to $10^{-5}$ \cite{DBLP:journals/corr/BowmanVVDJB15}, allowing the encoders to learn a feature set prior to increasing the weight on the KL term for the priors. 

To improve reconstruction performance, we also used a series of affine transformations to interpret the latent prior to decoding, as proposed in StyleGAN 2 \cite{DBLP:journals/corr/abs-1912-04958}.

To support a wide range of domains, our VAE supports inputs from multiple encoders and decoders.  This is particularly important for StarCraft II \cite{DBLP:journals/corr/abs-1708-04782} representations, which consist of multiple spatial semantic maps.  Each encoder can either be MLPs or convolutional, depending on the part of the input it captures.  The outputs of the encoders are concatenated and passed to a MLP for inferring the mean and standard deviations.  To improve the ability to reconstruct smaller details, convolutional encoders assemble a combined latent by constructing individual latents from each encoder layer.  

 For each environment, we used 95\% of the recorded trajectories for training and the remainder for testing.  Test mean-squared error for normalized predictions was under $0.1$ across the full range of $[-1, 1]$.

We now detail the three major experiments that form the core of our contributions.  The first compares gradient-driven counterfactuals across several RL environments.  The second examines how plausibility adjustments can improve the likelihood of a generated example and reduce the number of anomalous counterfactuals.  Finally, we demonstrate the effectiveness of the jointly training latent space on the quality of counterfactuals.
Equivalence to baselines from literature are marked when appropriate.

\subsection{Counterfactual Query Setup}
For each interestingness variable $i$ and sign of change $s$, we sampled 100 individual instances from the set of recorded trajectories.  Each instance $\langle\mathbf{x}_q, \mathbf{y}_q\rangle$ was filtered so there is sufficient margin in variable $i$ for a valid counterfactual, e.g., $-1 \leq \mathbf{y}_{q}^{(i)} + s \epsilon \leq 1$.  For the value function, $\epsilon$ was two times the standard deviation.  The other variables are in the range $[-1,1]$, and we set $\epsilon=0.5$.  From our inventory of three interestingness variables and two signs of change (increasing or decreasing their value), we experimented with a total of six combinations (600 queries) for each counterfactual generation method and environment.  

\subsection{Results}
We compared the following methods across the three RL environments:
\begin{itemize}
    \item Drawing the Nearest Unlike Neighbor from the training set (NUN)
    \item Latent Interpolation to the NUN, stopping at the first point where $\kappa_i$ is met (InterpPt)
    \item Using Iterative Gradient Update to perturb the latent until $\kappa_i$ is met (Gradient).
\end{itemize}




\begin{table}
\begin{tabular}{ L{37pt} R{70pt} R{55pt} R{25pt}}
\toprule
    \textbf{Method} &        \textbf{Obs Diff} &  \textbf{Anom Score} & \textbf{Valid CFs} \\
\\\\[-3.2\medskipamount]
    \multicolumn{4}{c}{\textbf{Cartpole}} \\
\hline\\\\[-3.5\medskipamount]
    \emph{NUN} & $1.28 \pm 0.71$ &  $0.11 \pm 0.03$ &        $\mathbf{1.00}$ \\
    \emph{InterpPt} & $\mathbf{0.86 \pm 0.82}$ &  $\mathbf{0.07\pm0.02}$ &        $0.50$ \\
    \emph{Gradient} & $0.99 \pm 0.89$ &  $0.31 \pm 0.54$ &          $0.98$ \\
\\\\[-3.2\medskipamount]
    \multicolumn{4}{c}{\textbf{Canniballs}} \\
\hline\\\\[-3.5\medskipamount]
    \emph{NUN} & $1754.91 \pm 89.31$ &   $\mathbf{0.00 \pm 0.00}$ &        $0.67$ \\
    \emph{InterpPt} &     $\mathbf{5.46 \pm 4.10}$ &   $0.20 \pm 0.57$ &        $0.67$ \\
    \emph{Gradient} &   $18.94 \pm 20.88$ & $12.56 \pm 17.13$ &        $\mathbf{0.99}$ \\
\\\\[-3.2\medskipamount]
    \multicolumn{4}{c}{\textbf{SC2 Assault}} \\
\hline\\\\[-3.5\medskipamount]
    \emph{NUN} & $1746.12 \pm 573.25$ &   $\mathbf{7.52 \pm 3.88}$ &        $\mathbf{1.00}$ \\
    \emph{InterpPt} & $1234.94 \pm 880.92$ & $30.38 \pm 24.87$ &        $0.67$ \\
    \emph{Gradient} &   $\mathbf{83.36 \pm 141.25}$ & $33.73 \pm 45.29$ &        $0.97$ \\
\bottomrule
\end{tabular}
\caption{Counterfactual methods for each environment, with microaveraged statistics across counterfactual quality measures.}

\label{tab:conf_cfs}
\end{table}
\begin{figure*}
    \centering
    \includegraphics[scale=0.6]{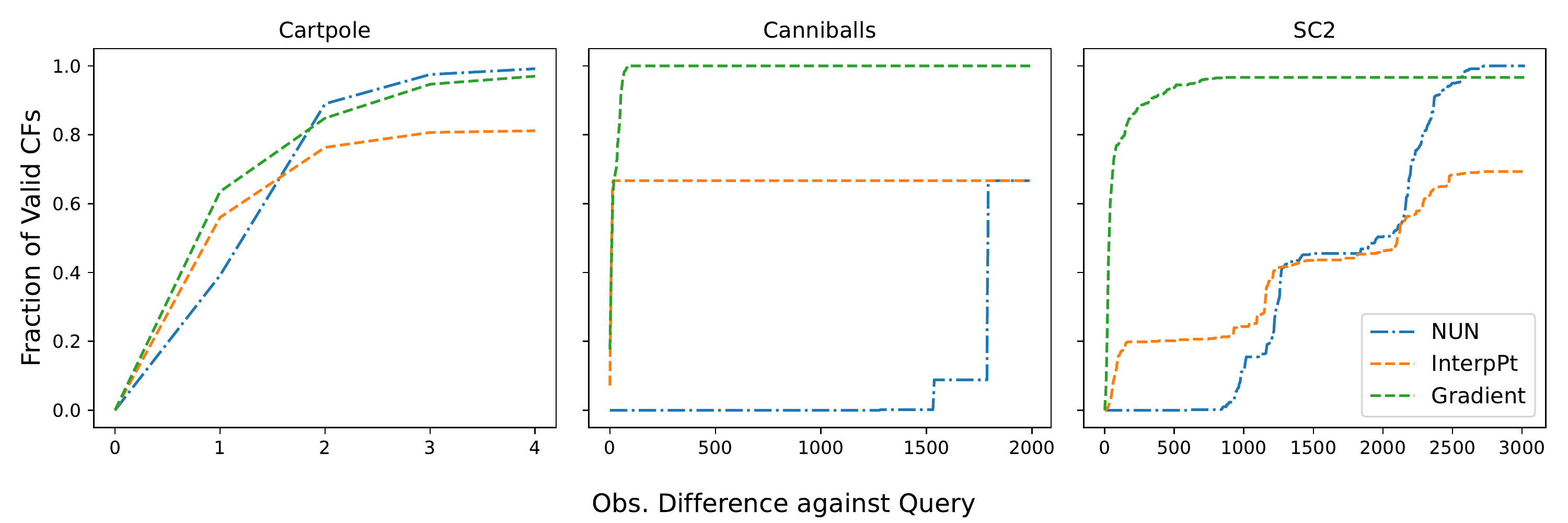}
    \caption{Number of valid counterfactuals (CFs, y-axis) at or below the  observational difference to the query (x-axis) for each environment.  In all environments, the latent-based approaches (Interpolation and Gradient) were able to produce counterfactuals not present in the memory (NUN) and that are more proximal to the query.}
    \label{fig:achieved_vs_threshold}
\end{figure*}

Table \ref{tab:conf_cfs} reports the micro-averaged mean and standard deviation of the observational differences, anomaly scores\footnote{Observational difference and anomaly score are the inverses of proximity and plausibility, so lower scores indicate better performance.},
and fraction of valid counterfactual queries for each method against the given query combinations for the three environments.  As expected, drawing from a memory of actual instances (NUN) produces the least anomalous and most plausible counterfactuals.  However, both latent-based approaches produce counterfactuals that are more proximal to the query instances, with significantly lower observational differences in Canniballs (InterpPt and Gradient) and SC2 Assault (Gradient).  These methods also feature overall lower mean observational differences while keeping anomaly scores within range of the NUN's scores.  

The Gradient method produced valid counterfactuals for most queries, missing at most $3\%$ overall.  The InterpPt method generated counterfactuals about $67\%$ of the time across all three environments, with the remainder requiring full traversal to the NUN.

\subsection{Proximity Analysis and Unseen Scenes}
In certain use cases, there is an upper bound on the distance between the query and a  valid counterfactual before they are deemed unrelatable \cite{keanegoodcounterfactuals2020}.  To assess this, we examined the proximity of valid counterfactuals to the query for each method.  Figure \ref{fig:achieved_vs_threshold} shows the cumulative fraction of valid counterfactuals whose observational differences from the query are at or below the given level, where ``valid'' is binary a binary pass/fail metric.  We find that the latent space methods (InterpPt and Gradient) were able to produce more valid counterfactuals compared to retrieving the closest counterfactual from memory (NUN).  These are also unseen scenes, as they would otherwise have been selected as counterfactuals from memory.  This effect is more pronounced in the Canniballs and SC2 Assault environments, as their observation spaces are significantly more complex than Cartpole's. 

\subsection{Impact of Plausibility Adjustment}
We examined the impact of a plausible scene gradient adjustment on counterfactuals for the SC2 Assault minigame environment.  
To arrive at a quantitative estimate of the number of actual anomalous scenes in the SC2 Assault minigame, we took 1392 pairs of real instances and sampled the reconstruction between them, giving us 4176 scenes\footnote{Earlier experimentation found anomalous scenes in between real instances to be extremely difficult to classify.}.  We then labeled these as anomalous or not, based on mistakes such as partially reconstructed units or duplicates of unique structures.  The 2196 training scenes were used to tune a threshold maximizing anomaly detection, giving a training accuracy of $96\%$ and a test accuracy of $95\%$ on the remaining 1980 test instances, over a baseline guess (plausible) of $66\%$.  The top half of Table \ref{tab:plausible_recon_table} shows the results, including the number of anomalous scenes generated.  Including the plausibility adjustments improves on both counterfactual generation methods, decreasing both the anomaly score and number of anomalies, with a slight cost in number of queries met for the interpolation methods.  


\begin{table*}
\centering
\begin{tabular}{l r r r r}
  \multicolumn{5}{c}{\textbf{Plausibility Adjustment Experiments}} \\
\toprule
           \textbf{Method} &       \textbf{Obs Diff} &  \textbf{Anomaly Score} & \textbf{Num Anom} & \textbf{Valid CFs} \\
\midrule
   \emph{InterpPt } & $1234.94 \pm 880.92$ & $30.38 \pm 24.87$ &  $4$ &        $0.67$ \\
   \emph{InterpPt no PlausAdj} & $1224.06 \pm 879.17$ & $48.26 \pm 75.52$ & $20$ &        $0.69$ \\
   \emph{Gradient} &   $83.36 \pm 141.25$ & $33.73 \pm 45.29$ & $46$ &        $0.97$ \\   
   \emph{Gradient no PlausAdj} &   $82.44 \pm 136.11$ & $34.41 \pm 46.62$ & $48$ &        $0.97$ \\
\bottomrule
    \\
      \multicolumn{5}{c}{\textbf{Experiments using Reconstruction-Only Latent}} \\
    \toprule
           \textbf{Method} &       \textbf{Obs Diff} &  \textbf{Anomaly Score} & \textbf{Num Anom} & \textbf{Valid CFs} \\
\midrule
        \emph{InterpPt} &  $881.94 \pm 770.84 $&   $72.00 \pm 70.78$ &  $69$ &        $0.44$ \\
        \emph{InterpPt no PlausAdj} &  $ 887.66 \pm 769.93$ & $135.76 \pm 146.80$ & $150$ &        $0.59$ \\
        \emph{Gradient} &  $124.17 \pm 100.85$ &   $83.07 \pm 44.34$ & $107$ &        $0.81$ \\
        \emph{Gradient no PlausAdj (xGEMs)}  &   $123.34 \pm 97.66$ &   $83.12 \pm 44.95$ & $107$ &        $0.81$ \\
        \bottomrule
\end{tabular}
\caption{Comparison of plausibility adjustments for counterfactual methods, showing that counterfactuals generated with a plausibility adjustment are more plausible, with a lower number of anomalous instances (top).  Counterfactual generation methods from a latent derived over a reconstruction-only generative model (bottom) show higher observational differences and more anomalous scenes in contrast with their counterparts that operated over a jointly trained latent space (top).}
\label{tab:plausible_recon_table}
\end{table*}

\subsection{Impact of Joint Training}

We test our hypothesis that joint training of the latent space with our desired outcome variables leads to better counterfactuals.  It is important to assess the benefits of jointly training input reconstruction and outcome prediction, as approaches in the literature trained these two tasks sequentially (see Sec. 2).  Using the SC2 Assault task, we trained the VAE model with just the reconstruction objective for 100 epochs, using the same training setup as the full model.  We then trained the outcome predictors given the latents produced by the reconstruction-only model.  This too was trained using the same setup, and these outcome predictors achieved prediction errors comparable to those of the full model.

We then re-ran the same set of experiments using the reconstruction-only latent and predictors.  For completeness, we included experiments without the plausibility adjustment.  Results are presented in the bottom half of Table \ref{tab:plausible_recon_table}.  Here we see that counterfactuals generated from the reconstruction-only latent space produced considerably more anomalous counterfactuals, both in terms of the score and in the number of generated counterfactuals that exceeded the anomaly threshold.  In addition, they produced fewer valid counterfactuals than their counterparts from the joint latent space.  We note that the Gradient approach without the plausbility adjustment over the Reconstruction-Only latent is equivalent to xGEMs \cite{joshixgems2018}, and it too would benefit from operating over a jointly trained latent.


\section{Discussion and Future Work}
We have presented two methods for obtaining counterfactuals from a generative model with a latent space jointly trained to reconstruct observations and predict outcome variables of interest. The first uses the latent space to create smooth interpolations between the query and case-based instances drawn from a memory, while the second uses gradient signals to iteratively update the query.  Both have been shown to produce previously unseen counterfactuals that are closer to the query.  We have shown that using a proxy signal for increasing the likelihood of a reconstruction can further improve the plausibility of counterfactuals from these methods.  Finally, we have shown that jointly training the generative model to both reconstruct and predict outcomes results in higher-quality counterfactuals.

Future areas of investigation include a closer examination of these methods in contrast to feature-level adversarial methods.  One of the major concerns from that class of methods is that the counterfactuals they generate may be imperceptible to humans, as their perturbations are fuzzing attacks that minimize feature-level changes.  In contrast, our latent space traversal methods have shown a perceptible amount of feature-level differences.  However, future work should include stronger assurances for preventing the generation of counterfactuals that are imperceptibly different from the query.

As we have seen, our gradient-based approach can improve the plausibility of counterfactuals generated by our methods.  This approach can be further improved to allow only adjustments that reflect certain types of feature-level edits that represent aspects of the agent or environment that a human operator has control over.  This is similar to the body of work around GAN steerability, where the output of a generative model is updated to reflect specific types of feature variation based off signals from trained discriminators \cite{harkonensteerability2020}.  

Finally, we note that this work, like many others, look at intrinsic measures of counterfactual quality.  Proper extrinsic evaluations of how counterfactuals can improve a meaningful task remains to be addressed.  One possible avenue would be to use counterfactuals to improve the quality of examples used for machine-teaching and tutoring applications \cite{simardmachineteaching2017}.  We are also investigating the use of directed counterfactuals to communicate likely or dangerous possible scenarios to decisionmakers ahead of time.  In addition, they may also act as a source of additional weak evidence for observational assessments of the causal link between features and outcomes.




\section{Acknowledgments}
This material is based upon work supported by the Defense Advanced Research Projects Agency (DARPA) under Contract No. HR001119C0112. Any opinions, findings and conclusions or recommendations expressed in this material are those of the author(s) and do not necessarily reflect the views of the DARPA.

\bibliographystyle{aaai}
\bibliography{main}

\end{document}